\documentclass[transmag,letterpaper]{IEEEtran}
\usepackage{nopageno}
\newif\ifpeerreview

\peerreviewfalse

\newcommand{\paperID}{XXXX}

\usepackage[switch]{lineno}

\usepackage{cite}
\usepackage{url}
\usepackage{lipsum}
\usepackage{times}
\usepackage{graphicx}
\usepackage{amsmath}
\usepackage{amssymb}
\usepackage{color}
\usepackage{multirow}
\usepackage{hhline}
\usepackage{animate}
\ifCLASSINFOpdf
\else
\fi
\begin{document}

\ifpeerreview
  \linenumbers
  \linenumbersep 5pt\relax
\fi 

\title{A Bit Too Much? High Speed Imaging from\\ Sparse Photon Counts}

\ifpeerreview
\author{Anonymous ICCP 2019 submission \\
Paper ID \paperID}
\else
\author{\IEEEauthorblockN{Paramanand Chandramouli\IEEEauthorrefmark{1},
Samuel Burri\IEEEauthorrefmark{2},
Claudio Bruschini\IEEEauthorrefmark{2}, 
Edoardo Charbon\IEEEauthorrefmark{2}, and
Andreas Kolb\IEEEauthorrefmark{1}}
\IEEEauthorblockA{\IEEEauthorrefmark{1}University of Siegen, Germany}
\IEEEauthorblockA{\IEEEauthorrefmark{2}Swiss Federal Institute of Technology (EPFL), Neuch\^atel, Switzerland}
}
\fi
\ifpeerreview
\markboth{Anonymous ICCP 2019 submission ID \paperID}%
{}
\else
\fi

\IEEEtitleabstractindextext{%
\begin{abstract}
Recent advances in photographic sensing technologies have made it possible to achieve light detection in terms of a single photon. Photon counting sensors are being increasingly used in many diverse applications. We address the problem of jointly recovering spatial and temporal scene radiance from very few photon counts. Our ConvNet-based scheme effectively combines spatial and temporal information present in measurements to reduce noise. We demonstrate that using our method one can acquire videos at a high frame rate and still achieve good quality signal-to-noise ratio. Experiments show that the proposed scheme performs quite well in different challenging scenarios while the existing approaches are unable to handle them.
\end{abstract}

\ifpeerreview
\else
\begin{IEEEkeywords}
Photon Counting Sensors, SPAD Imaging, Convolutional Neural Network, High Speed Imaging
\end{IEEEkeywords}
\fi
}

\maketitle
\thispagestyle{empty}

\IEEEdisplaynontitleabstractindextext

\section{Introduction}
\begin{figure*}[htb]
\begin{center}
\begin{tabular}{ccc}
\includegraphics[width=162pt]{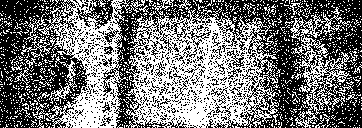}&\hspace{-12pt}
\includegraphics[width=162pt]{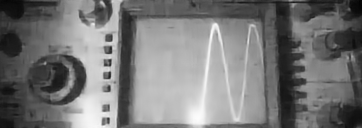}&\hspace{-12pt}
\includegraphics[width=162pt]{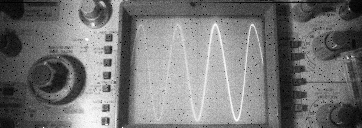}\\
(a)&\hspace{-12pt}(b)&\hspace{-12pt}(c)\\
\end{tabular}
\caption{Imaging wave propagation in an oscilloscope: (a) A 1-bit frame from the input to our algorithm captured at 156000 fps. (b) Corresponding resultant frame of the proposed scheme. (c) Average of 255 frames with the frame shown in (a) as the center (the dark pixels are due to sensor defects). Kindly refer to the supplementary material for the complete video. \label{fig:osc_intro}}
\end{center}
\end{figure*}

We address the problem of imaging fast dynamic scenes through single photon counting sensors \cite{fossum2018photon}. Single photon avalanche diode (SPAD) detectors are endowed with the ability of photon counting and time-stamping. They are getting increasingly popular for a variety of imaging applications  \cite{charbon2014single,shin2016photon,gariepy2015single,o2017reconstructing,heide2017robust}. We attempt to the enhance the fast motion capture capability of SPAD sensors by developing a robust video recovery algorithm. Consider the example shown in Fig. \ref{fig:osc_intro}. An analog oscilloscope with a traversing sinusoid was imaged by a SwissSPAD camera \cite{burri2014architecture}. Fig. \ref{fig:osc_intro} (a), shows a frame from the captured video obtained at 156k frames per second (fps). The value of each pixel in this frame is a binary number wherein positive values indicate the detection of a photon. Note that in a typical DSLR camera, the number of photons captured is of the order of thousands of photons per pixel \cite{nakamura2017image,shin2016photon} while the frame in Fig. \ref{fig:osc_intro} (a) shows the detection of \emph{only one photon}. Due to the extremely low photon count, one can hardly infer any structure present in the scene. One could average consecutive frames to reduce the effect of noise while sacrificing the temporal resolution (Fig. \ref{fig:osc_intro} (c)). To preserve temporal resolution as well as recover accurate scene reflectance, we develop a convolutional neural network (CNN) that takes the low-photon count sequence as input and generates a high-photon count estimate at the \emph{same frame rate}. Our scheme effectively combines the spatio-temporal information present in the input sequence for video recovery. The resultant frame from our method is shown in Fig. \ref{fig:osc_intro} (b). Note that in Figs \ref{fig:osc_intro} (a) and (b), one can observe the localization of the sinusoidal wave while in Fig. \ref{fig:osc_intro} (c), the temporal information is lost. In Fig. \ref{fig:osc_intro} (b), we also observe that the details of the static regions have been recovered quite well.


Previously, in the context of time-correlated SPAD imaging, regularization-based approaches have been developed to reconstruct scene reflectivity \cite{shin2016photon,yan2017photon}. For oversampled binary observations, image reconstruction algorithms such as \cite{s16111961} are applicable. In this paper, our objective is to jointly recover spatial and temporal variations of radiance in dynamic scenes without spatial oversampling. Existing video denoising schemes devise methods for combining local and non-local structures present across space and time \cite{Maggioni,sutour2014adaptive,Wen_2017_ICCV}. In extremely noisy scenarios, explicitly determining such information does not work well. Instead of ``hand-crafted'' approaches to combine structural information, our method uses convolutional neural networks consisting of 3D filtering across the spatio-temporal volume \cite{tran2015learning}. By accumulating a set of binary frames, one can obtain video sequences with lesser noise and reduced frame rate. In this paper, we address different scenarios in which different number of the binary frames are combined. Although we show the application of our method on SwissSPAD cameras, our scheme can be applied to any other photon counting or binary imaging sensors \cite{Dutton,gyongy2016256}. High speed consumer cameras typically require significantly bright illumination \cite{web:illum}. In contrast, we do not use any high intensity illumination and operate in normal lighting conditions.

\subsection{Related work} Since our work is related to SPAD imaging, photon counting sensors and video denoising, we briefly discuss relevant prior works in these topics.

\textbf{SPAD-based imaging} SPAD sensors are photodetectors in which photon radiation can be detected from the resulting large avalanche currents. SPAD sensor arrays are capable of photon counting at a high speed with high timing resolution and are useful in a variety of applications such as fluorescence lifetime imaging microscopy (FLIM), positron emission tomography (PET), time-of-flight imaging etc. \cite{shin2016photon,charbon2014single,li2010real}. Recently, Burri et al. developed a SPAD array known as SwissSPAD \cite{burri2014architecture}. The SwissSPAD is fabricated in a high-voltage CMOS process and features a large 512$\times$128 array with global gating. In this paper, we use data from different SwissSPAD sensor arrays for demonstrating our high speed video recovery scheme.

Recent works in range imaging through SPAD sensors include \cite{kirmani2014first,shin2016photon,lindell2018single}. SPAD cameras have been used to perform challenging tasks such as transient imaging \cite{o2017reconstructing,gariepy2015single,lindell2018towards,sun2018depth,pediredla2018signal}, non-line-of sight imaging \cite{buttafava2015non,gariepy2016detection,heide2017robust,o2018confocal} and imaging through fog \cite{satat2018towards}.

\textbf{Quanta Imaging} Closely related to the topic of single photon counting is that of Binary/Quanta image sensors \cite{fossum2005sub,fossum2018photon}. These sensors were developed with the objective of shrinking the pixel pitch. In Quanta imaging, the densely defined sensors oversample the scene radiance to generate binary measurements \cite{fossum2005sub,Yang}. Image reconstruction schemes have been proposed for these sensors \cite{s16111961,choi2018image,remez2016picture,rojas2017learning}. The main difference between these reconstruction methods and our scheme is that these algorithms consider the availability of a higher number of samples to estimate the image intensity at a pixel location. These methods have mainly focused on recovering a static image. In contrast, our focus is on recovering videos.

\textbf{Video denoising}
Video denoising has been widely studied for many years and different kinds of algorithms have been proposed. Because of the vastness of the topic, we restrict our discussion to some of the popular and recent works.  When compared with applying image denoising on each frame, video denoising has an advantage because the high temporal coherence can be leveraged to make a better prediction. Consequently, denoising algorithms adopt different strategies such as non-local means \cite{buades2005denoising,sutour2014adaptive}, motion estimation \cite{liu2010high,werlberger2011optical}, 3D dictionary representations \cite{protter2009image} etc. Maggioni et al. \cite{Maggioni} propose a denoising method popularly known as VBM4D and is based on the collaborative filtering scheme (VBM3D) \cite{Dabov2007VideoDB}. They apply this filtering scheme to a stack of 3-D spatio-temporal volumes that are obtained through non-local grouping. Sutour et al., \cite{sutour2014adaptive} develop a variational denoising scheme by adaptively combining nonlocal means with total variation on spatio-temporal volumes. Their method adapts to different noise levels. The authors in \cite{ji2010robust} propose a denoising method for the scenario of mixed noise (like Gaussian noise mixed with impulsive noise). They formulate the denoising problem as low-rank matrix completion problem. This work has been extended in \cite{ji2011robust} to avoid pre-detection of outliers. 

Recently, techniques based on deep learning have been proposed for video denoising. Chen et al. \cite{chen2016deep} propose a deep recurrent neural network for video denoising. Their method does not assume a specific noise model and performs close to state-of-the art VBM4D scheme \cite{Maggioni}. Xue et al. \cite{Xue2017VideoEW} propose task oriented flow to achieve a specific video processing objective such as denoising. Instead of trying to achieve a precise flow estimation, they train a model whose objective is to predict a motion field tailored for a specific task. In our experiments, we observe that video denoising methods fail in the scenario of SPAD imaging. This could be due to the fact that grouping of similar structures across the spatio-temporal volume fails when the number of photons in the observation is sparse.

\textbf{Contributions} The contributions of this paper can be summarized as follows: i) We address the problem of joint spatio-temporal radiance estimation from single photon counting sensors. ii) We develop a CNN-based video recovery scheme that can handle very high noise levels and still maintain the high frame rate iii) We devise methods to obtain appropriate real and synthetic datasets for training and evaluation. This data will be made available subsequently.
\section{Image formation}
In this section, we describe the image formation process for SwissSPAD cameras which have a globally gated sensor array \cite{burri2014architecture}. The imaging model can also be applied to any other gated SPAD arrays \cite{Dutton,gyongy2016256}.

Each pixel in the SwissSPAD array is composed of a SPAD p-n junction suitably biased for enabling photon triggered avalanche. A one-bit counter is present at every pixel. The SPAD array has a global shutter in which all the pixels can be kept active for a duration as low as 3.8 ns. The pixel counter content is transferred from the sensor pixels via a fast digital readout which takes about 6.4 $\mu$s for transferring the contents of the whole array of $128{\times}512$ pixels. In effect, a 1-bit frame of size 128{$\times$}512 indicating the detection of \emph{one} photon will be read out in 6.4 $\mu$s resulting in a frame rate of 156kfps \cite{burri2014architecture}.

Due to the dark counts generated in a SPAD by thermal events, the number of photon counts per unit time follows a Poisson distribution \cite{antolovic2016nonuniformity}. The probability of $k$ photon counts in unit time is given by 
\begin{align}
p_c\left(k\right) = \frac{{\chi}^ke^{-{\chi}}}{k!}
\end{align}
where $\chi$ is the expected value of counts and is related to the impinging count rate, dark count rate and photon detection efficiency \cite{antolovic2016nonuniformity}. Within a particular time frame, the SwissSPAD sensor array can only report whether one or more photons were detected. Consequently, the probability of recording a detection in one readout time is $P\left(\mbox{count}>0\right) = 1-e^{-{\chi}}$. Since the number of photons impinging at a pixel depends on the scene radiance corresponding to that pixel, one can conclude that the scene radiance is sensed non-linearly (according to $1-e^{-{\chi}}$). This non-linear mapping has been experimentally verified in \cite{antolovic2016nonuniformity}.

\begin{figure*}
\begin{center}
\begin{tabular}{ccccc}
\includegraphics[width=84pt,height = 100pt]{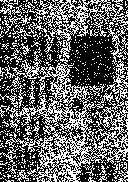}&\hspace{-2pt}
\includegraphics[width=84pt,height = 100pt]{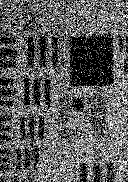}&\hspace{-2pt}
\includegraphics[width=84pt,height = 100pt]{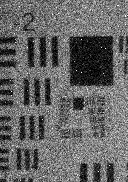}&\hspace{-2pt}
\includegraphics[width=84pt,height = 100pt]{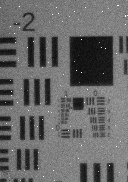}&\hspace{-2pt}
\includegraphics[width=84pt,height = 100pt]{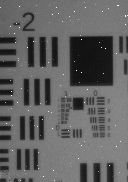}\\
(a)&\hspace{-2pt}(b)&\hspace{-2pt}(c)&\hspace{-2pt}(d)&\hspace{-2pt}(e)
\end{tabular}
\caption{Resolution chart imaged by the SwissSPAD camera at different bits: (a) 1-bit frame. (b) 2-bit frame. (c) 4-bit frame. (d) 8-bit frame. (e) 14-bit frame. \label{fig:im_model}}
\end{center}
\end{figure*}

Since the SwissSPAD camera records only a `binary pixel' in each frame, at a time instant $t$, the intensity $I_t(i,j)$ observed at a pixel $(i,j)$ is a Bernoulli sample whose probability depends on the intensity of the corresponding scene point. Fig. \ref{fig:im_model} (a) shows a single 1-bit image of a resolution chart obtained by the SwissSPAD camera. When $3$ such frames were captured and averaged, the resultant image is shown in Fig. \ref{fig:im_model} (b). This is a 2-bit image corresponding to $2^2-1$ frames. Similarly Figs. \ref{fig:im_model} (c), (d) and (e) show 4-bit, 8-bit and 14-bit images. As the number of samples are increased, the noise in the observation reduces. The isolated bright pixels present in all the observations of Fig. \ref{fig:im_model} correspond to the hot pixels.


\subsection{Objective}
When a binary image sequence $I_t$ is averaged upto $b$ bit levels, one would get another sequence denoted by $u_{\tau}^b$ with frame rate reduced by the factor $N_b=2^b-1$. The sequence with bit resolution $b$ is given by 
\begin{align}
u^b_{\tau} = \frac{1}{N_b}\sum_{t=0}^{N_b-1}I_{t+{\tau}N_b}
\end{align}

 While imaging a static scene, one can afford to collect as many frames as possible and average them for obtaining an accurate estimate of the scene reflectance map. However, as seen in Fig. \ref{fig:osc_intro}, while imaging fast dynamic scenes, if one were to average many frames, the temporal information would be lost. Our objective is to overcome this trade-off between frame rate and intensity resolution. For a particular scene, consider that one requires to have a specific frame rate and thereby the bit resolution $b$ gets fixed to a certain level. Let $u^b_{\tau}$ denote the corresponding sequence. Our aim is to estimate a high bit intensity sequence $u^{\tilde{b}}_{\tau}$ where $\tilde{b}$ is much greater than $b$ and also at the same time preserve the frame rate of the original sequence.

 In this paper, we consider the scenarios where $N_b$ takes the values of either $1$, $3$, $7$, and $15$. i.e., either 1-bit, 2-bit, 3-bit or 4-bit input sequences, respectively. From image sequences of such low photon counts, we attempt to recover sequences corresponding to a very high bit resolution $\tilde{b}>12$ bits. Note that, beyond a certain value of number of bits, any further increase would not be adding new information \cite{fossum2016quanta}. We observed that when $\tilde{b}>12$, the noise becomes imperceptible.
 



\section{Proposed method}
We propose to learn a mapping function $f$ for generating a sequence with a high bit resolution $\hat{u} = f\left(u^b;{\mathbf{\theta}}\right)$, such that it is close to the noise-free sequence $u^{\tilde{b}}$. Note that we have dropped the time index $\tau$ to simplify the notation. The term $\mathbf{\theta}$ denotes the network parameters.
\subsection{Network architecture}
\begin{figure}
\begin{center}
\begin{tabular}{c}
\includegraphics[width=220pt]{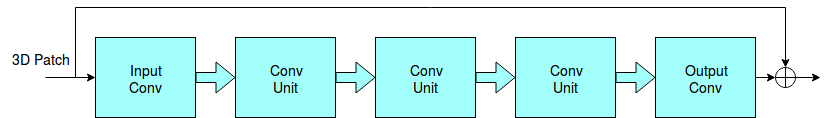}
\end{tabular}
\caption{One residual unit from the proposed ConvNet architecture. Overall, the network consists of $K$ such cascaded residual blocks. \label{fig:net}}
\end{center}
\end{figure}
Our network architecture is composed of 3D convolutional layers and residual blocks. The network design is motivated by the fact that many image restoration methods employ architectures with a similar structure (but with 2D convolutions) \cite{lai2017deep,fu2017removing,jiao2017formresnet}. Following such an approach also helps to avoid gradient exploding/vanishing problems \cite{he2016deep}. Since 2D CNNs in ResNet style have achieved significant success, we intend to use 3D convolutional layers for processing videos. Fig. \ref{fig:net} shows one residual block of our network. Totally, our network consists of $K=3$ such residual blocks. Each residual block, has units that are composed of 3D convolutional filters of size $3{\times}3{\times}3$. The input convolutional unit consists of one input and $60$ output channels. While the three intermediate `conv' units consist of $60$ input and $60$ output channels, the output unit has $1$ output channel. Except the output unit, all the others are followed by a `Leaky' ReLU to model non-linearities. For comparison, we also train with filter size $5{\times}5{\times}5$.

 The input to the network is a spatio-temporal patch. At the end of each residual block indexed by $k$, the input is added to the resultant of the `output conv' layer to arrive at $\hat{u}_k$, an estimate of the spatio-temporal patch corresponding to the clean high-bit sequence. For training the network, we minimize the loss function which is composed of the loss functions of each residual block. Since it is observed in \cite{lai2017deep} that the Charbonnier penalty function leads to good performance as against the standard $\ell_2$ penalty, we also use the Charbonnier penalty to define our loss function. i.e., the final loss function is given by
 \begin{align}
 E\left(\hat{u}_k,u^{\tilde{b}},\mathbf{\theta}\right) = \frac{1}{N}\sum_{i=1}^N\sum_{k=1}^K\rho\left(\hat{u}_k-u^{\tilde{b}}\right)
 \end{align}
 where $N$ is the number of training samples in a batch and the penalty term is given by $\rho(x) = \sqrt{x^2+{\eta}^2}$. The value of $\eta$ is chosen to be $10^{-3}$ following \cite{lai2017deep}.

\subsection{Training data}
\label{sec:data_gen}
For each bit-level, we train our model using simulated data. For the case of 4-bit sequences, we also train with real pairs of low-bit and corresponding high-bit sequences obtained from SPAD camera. For the 1-bit and 2-bit scenarios, the frame rates that can be achieved by SwissSPAD are 156k fps and 78k fps, respectively. At such high frame rates, one can expect that the temporal variations are quite low. Hence, for these two scenarios, we generate a video dataset with high temporal coherence. The raw videos used in the training of video-deblurring scheme of \cite{su2017deep}, consists of image sequences at 240fps. The spatial extent of these sequences is $1280{\times}720$. We spatially downsample these sequences by a factor of $7$ to obtain sequences with reduced variations across time. We randomly crop these downsampled sequences at different temporal locations and arrive at $2500$ sequences of dimension $100{\times}100{\times}64$. These sequences are directly considered to be the high intensity resolution sequences ($u^{\tilde{b}}$). To generate the low-bit sequences, for every frame, we average $N_b$ Bernoulli sampled (binary) instances. In each Bernoulli instance, the probability of getting a one at a pixel is equal to the corresponding normalized true intensity value at that pixel. i.e., brighter pixels are more likely to generate a $1$. To generate such a sample, we generate a uniform random number at every pixel. At a particular pixel, if the randomly generated number is less than the true normalized image intensity value, then that pixel is assigned as one, and zero otherwise. We also randomly add salt-and-pepper-noise at a few points to simulate hot-pixels. For 3-bit and 4-bit scenarios, we found that training with videos from UCF 101 dataset \cite{soomro2012ucf101} that have a normal frame-rate was sufficient. We explain the procedure of obtaining real training data from the SPAD camera in the supplementary material.

\textbf{Model training}
We trained our network models on a NVIDIA GeForce GTX 1080 Ti using stochastic gradient descent without batch normalization. The models named 3DCNN1B, 3DCNN2B, 3DCNN3B and 3DCNN4B denote the CNNs trained with 1-bit, 2-bit, 3-bit and 4-bit sequences, respectively. The network trained using 4-bit real data is referred to as 3DCNNR. From the $2500$ sequences, we use $2400$ for training and the rest for testing. The input in each batch of training is obtained by randomly cropping 3D patches of size $60{\times}60{\times}38$ from any of the sequences of the training dataset. We randomly resize, flip, and rotate by multiples of $90^{\circ}$ for data augmentation.
%
\begin{table}[htb]
	\renewcommand{\arraystretch}{1.1}
	\caption{Quantitative evaluation on simulated data.}
	\label{table_12b}
	\begin{center}
\resizebox{\columnwidth}{!}{%
		\begin{tabular}{|c|| c|c|c|c|}
			\hline
			

{\bfseries Measure} & 3DCNN1B & 3DCNN1B (5) & 3DCNN2B & 3DCNN2B (5)\\
			\hline

			\ PSNR & 25.44 & 25.37 & 27.40 & 26.86 \\
			\hline
			\ SSIM & 0.744 & 0.740& 0.815 & 0.803\\
			
			\hline
		\end{tabular}
}
	\end{center}
\end{table}
\begin{figure*}[htb]
\begin{center}
\begin{tabular}{ccccc}
\includegraphics[width=92pt]{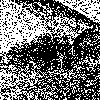}&\hspace{-12pt}
\includegraphics[width=92pt]{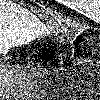}&\hspace{-12pt}
\includegraphics[width=92pt]{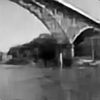}&\hspace{-12pt}
\includegraphics[width=92pt]{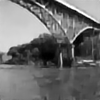}&\hspace{-12pt}
\includegraphics[width=92pt]{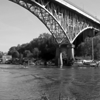}\\
\includegraphics[width=92pt]{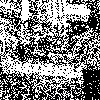}&\hspace{-12pt}
\includegraphics[width=92pt]{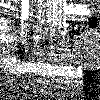}&\hspace{-12pt}
\includegraphics[width=92pt]{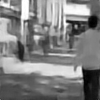}&\hspace{-12pt}
\includegraphics[width=92pt]{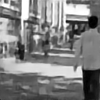}&\hspace{-12pt}
\includegraphics[width=92pt]{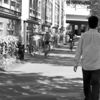}\\
\includegraphics[width=92pt]{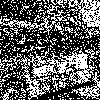}&\hspace{-12pt}
\includegraphics[width=92pt]{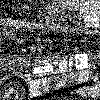}&\hspace{-12pt}
\includegraphics[width=92pt]{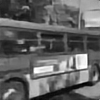}&\hspace{-12pt}
\includegraphics[width=92pt]{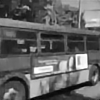}&\hspace{-12pt}
\includegraphics[width=92pt]{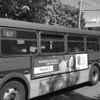}\\
\includegraphics[width=92pt]{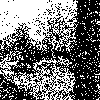}&\hspace{-12pt}
\includegraphics[width=92pt]{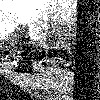}&\hspace{-12pt}
\includegraphics[width=92pt]{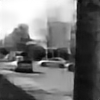}&\hspace{-12pt}
\includegraphics[width=92pt]{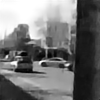}&\hspace{-12pt}
\includegraphics[width=92pt]{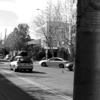}\\
\includegraphics[width=92pt]{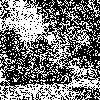}&\hspace{-12pt}
\includegraphics[width=92pt]{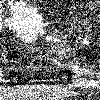}&\hspace{-12pt}
\includegraphics[width=92pt]{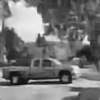}&\hspace{-12pt}
\includegraphics[width=92pt]{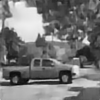}&\hspace{-12pt}
\includegraphics[width=92pt]{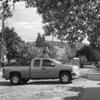}\\
One-bit frame&\hspace{-12pt}Two-bit frame&\hspace{-12pt}3DCNN1B result&\hspace{-12pt}3DCNN2B result&\hspace{-12pt}Ground truth
\end{tabular}
\caption{Representative examples from our test set (Videos are in supplementary material)\label{fig:1b2bs}}
\end{center}
\end{figure*}

\begin{figure*}[htb]
\begin{center}
\begin{tabular}{ccc}
\includegraphics[width=162pt]{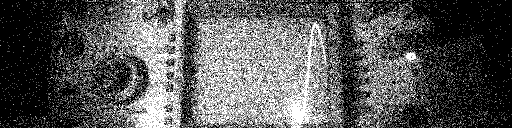}&\hspace{-12pt}
\includegraphics[width=162pt]{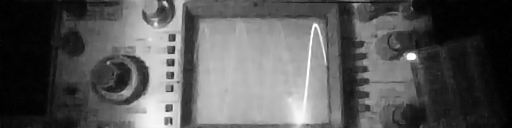}&\hspace{-12pt}
\includegraphics[width=162pt]{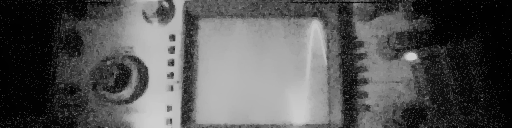}\\
(a)&\hspace{-12pt}(b)&\hspace{-12pt}(c)\\
\includegraphics[width=162pt]{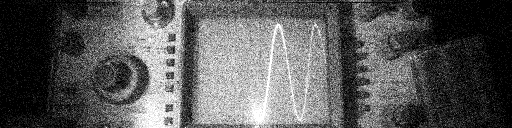}&\hspace{-12pt}
\includegraphics[width=162pt]{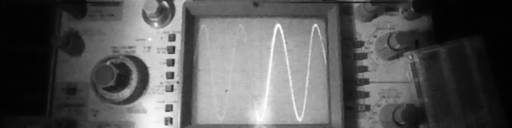}&\hspace{-12pt}
\includegraphics[width=162pt]{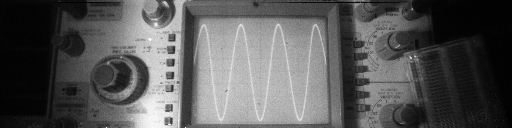}\\
(d)&\hspace{-12pt}(e)&\hspace{-12pt}(f)\\
\end{tabular}
\caption{(a) A 2-bit frame from the input to our algorithm captured at 78000 fps, and (b) corresponding resultant frame (3DCNN2B). (c) Output of \cite{sutour2014adaptive}. (d) A 4-bit frame, and (e) corresponding output from 3DCNNR. (f) Average of 120 4-bit frames. \label{fig:osci2b4b}}
\end{center}
\end{figure*}

\begin{figure}[htb]
\begin{center}
\begin{tabular}{cc}
\includegraphics[width=62pt]{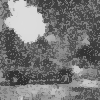}&\hspace{-12pt}
\includegraphics[width=62pt]{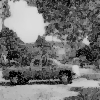}\\
\end{tabular}
\caption{Result of the algorithm in \cite{sutour2014adaptive} for the input shown in the last row of Fig. \ref{fig:1b2bs}: Output for (left) 1-bit sequence and (right) two-bit sequence. \label{fig:rnl_mb}}
\end{center}
\end{figure}

\section{Experiments}
To run our algorithm on a GPU for videos of realistic size, we divide an input sequence into overlapping 3D spatio-temporal patches and subsequently merge the outputs. For sequences of a particular bit-level, we use its corresponding CNN. If the bit-level of an input is not known, it can be determined easily by checking the number of unique levels in the pixel intensities.

\noindent \textbf{1-bit and 2-bit synthetic experiments} We used the test set of simulated data to evaluate the performance of our video reconstruction method. Table \ref{table_12b} shows the peak signal to noise ratio (PSNR) and structural similar index measure (SSIM) averaged over all the $100$ test sequences. Note that 3DCNN1B and 3DCNN2B had different inputs, but the ground-truth sequences were the same as seen in the representative examples shown in Fig. \ref{fig:1b2bs}. In Table \ref{table_12b}, the value $5$ within parentheses indicates that the filter size used in the CNN was $5{\times}5{\times}5$ instead of $3{\times}3{\times}3$. To check if any other denoising method works for this scenario, we applied the algorithms proposed in \cite{Maggioni}, \cite{sutour2014adaptive}, \cite{Xue2017VideoEW}, and \cite{denoisell} on five of these image sequences. None of these methods were able to restore videos for these sequences. We varied the parameters of the algorithms and searched for the optimal values. Out of these other algorithms, the best performance was obtained from \cite{sutour2014adaptive} (when applied with Poisson noise statistics). However even this is not quite satisfactory. The SSIM values of the outputs from \cite{sutour2014adaptive} ranged between $0.5$ and $0.64$. For visual comparison, we show one example output of \cite{sutour2014adaptive} in Fig. \ref{fig:rnl_mb}. The synthetic experiments clearly demonstrate that existing video denoising methods cannot be used for observations with sparse photon counts. However, we subsequently notice improvements in their performance when the bit-level improves.

We also checked if a single network can be used to recover both 1-bit and 2-bit sequences. For this purpose, we trained another CNN wherein the inputs for training consisted of both 1-bit and 2-bit sequences (with equal probability). We observe a slight reduction in the performance of this jointly-trained CNN compared to that of the specific networks seen in Table \ref{table_12b}. For the case of 1-bit test sequences, the resultant mean PSNR and SSIM from the jointly-trained CNN are $24.97$ and $0.726$, respectively. For the 2-bit test sequences, the resultant mean PSNR and SSIM are $26.9$ and $0.80$, respectively. We have also tested the proposed networks on video sequences with downsampling factors less than $7$. In these test sequences, the spatio-temporal variations will be increased when compared to the training data. These results are reported in the supplementary material.

\begin{figure}[htb]
\begin{center}
\begin{tabular}{cc}
\includegraphics[width=112pt,height = 54pt]{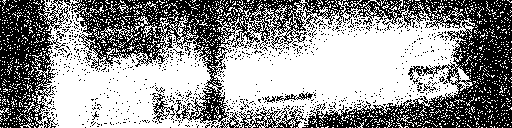}&\hspace{-12pt}
\includegraphics[width=112pt,height = 54pt]{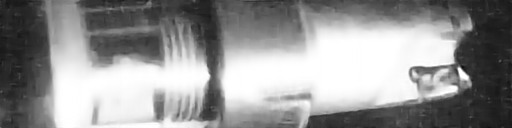}\\
(a)&\hspace{-12pt}(b)\\
\includegraphics[width=112pt,height = 54pt]{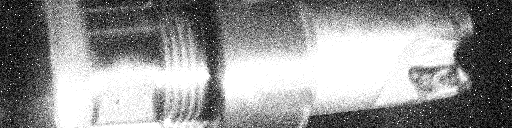}&\hspace{-12pt}
\includegraphics[width=112pt,height = 54pt]{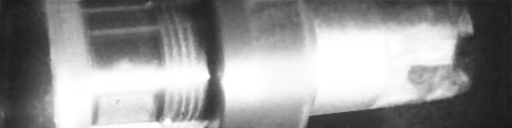}\\
(c)&\hspace{-12pt}(d)\\
\end{tabular}
\caption{High speed rotating tool (1-bit): (a) A frame from the input sequence and (b) its corresponding output from 3DCNN1B. (c) A 4-bit input frame generated by averaging. (d) Output of 3DCNN4B on the 4-bit input.\label{fig:b1_blade}}
\end{center}
\end{figure}

\begin{figure}[htb]
\begin{center}
\begin{tabular}{cc}
\includegraphics[width=112pt,height = 54pt]{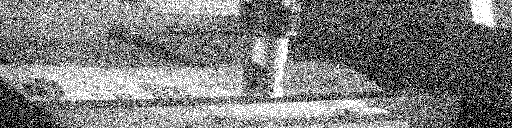}&\hspace{-12pt}
\includegraphics[width=112pt,height = 54pt]{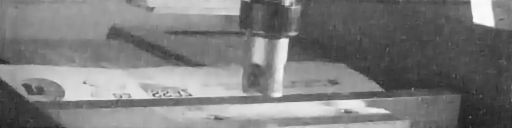}\\
(a)&\hspace{-12pt}(b)\\
\includegraphics[width=112pt,height = 54pt]{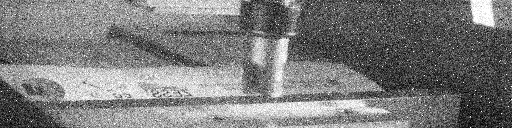}&\hspace{-12pt}
\includegraphics[width=112pt,height = 54pt]{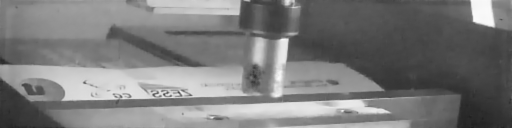}\\
(c)&\hspace{-12pt}(d)\\
\end{tabular}
\caption{High speed rotating tool (2-bit): (a) A frame from the input sequence and (b) its corresponding output from 3DCNN2B. (c) A 4-bit input frame generated by averaging. (d) Output of 3DCNN4B on the 4-bit input.\label{fig:b2}}
\end{center}
\end{figure}

\begin{figure*}[htb]
\begin{center}
\begin{tabular}{cccccc}
\includegraphics[width=64pt,height = 150pt]{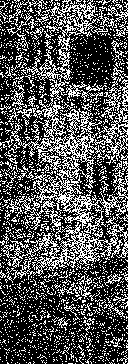}&\hspace{-12pt}
\includegraphics[width=64pt,height = 150pt]{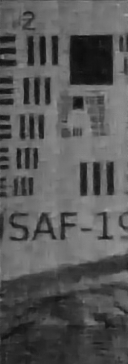}&\hspace{-12pt}
\includegraphics[width=64pt,height = 150pt]{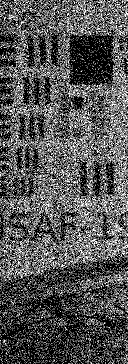}&
\hspace{-12pt}
\includegraphics[width=64pt,height = 150pt]{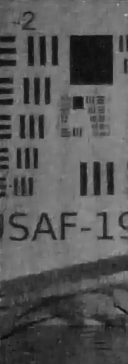}&\hspace{-12pt}
\includegraphics[width=64pt,height = 150pt]{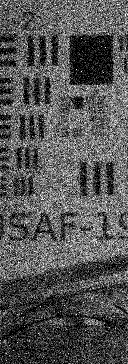}&\hspace{-12pt}
\includegraphics[width=64pt,height = 150pt]{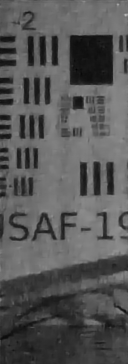}\\
\includegraphics[width=64pt,height = 150pt]{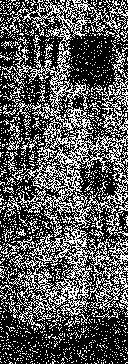}&\hspace{-12pt}
\includegraphics[width=64pt,height = 150pt]{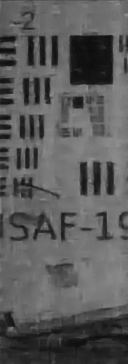}&\hspace{-12pt}
\includegraphics[width=64pt,height = 150pt]{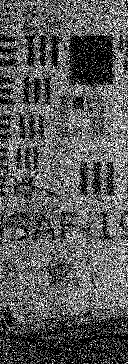}&\hspace{-12pt}
\includegraphics[width=64pt,height = 150pt]{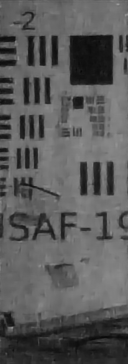}&\hspace{-12pt}
\includegraphics[width=64pt,height = 150pt]{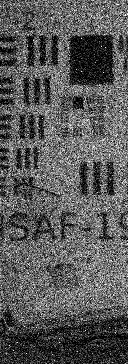}&\hspace{-12pt}
\includegraphics[width=64pt,height = 150pt]{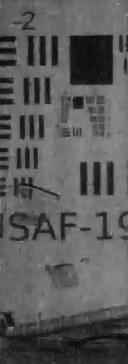}\\
\end{tabular}
\caption{Each row indicates a different time instant. (Left Pair) 1-bit observation and result. (Central) 2-bit observation and result. (Right) 3-bit observation and result.\label{fig:glass}}
\end{center}
\end{figure*}

\begin{figure}[htb]
\begin{center}
\begin{tabular}{cc}
\includegraphics[width=112pt,height = 54pt]{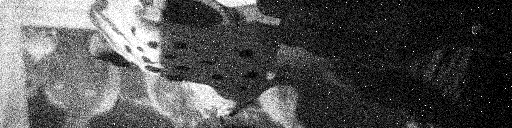}&\hspace{-12pt}
\includegraphics[width=112pt,height = 54pt]{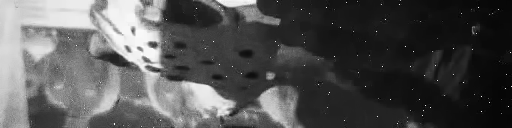}\\
(a)&\hspace{-12pt}(b)\\
\includegraphics[width=112pt,height = 54pt]{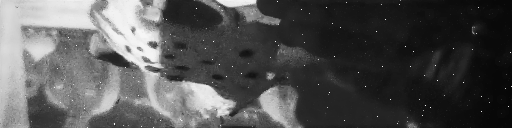}&\hspace{-12pt}
\includegraphics[width=112pt,height = 54pt]{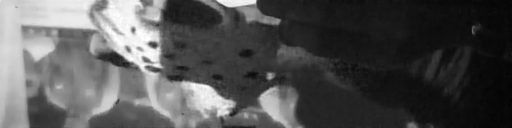}\\
(c)&\hspace{-12pt}(d)\\
\end{tabular}
\caption{Real example of a balloon bursting sequence. Frame from (a) input sequence, result of (b)  \cite{Maggioni}, (c) \cite{sutour2014adaptive}, (d) 3DCNN4B. \label{fig:balloon}}
\end{center}
\end{figure}

\begin{figure*}[htb!!!]
\begin{center}
\begin{tabular}{ccc}
\includegraphics[width=162pt]{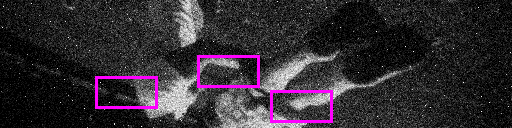}&\hspace{-12pt}
\includegraphics[width=162pt]{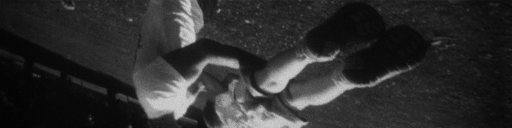}&\hspace{-12pt}
\includegraphics[width=162pt]{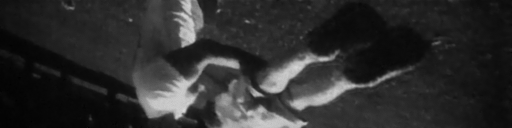}\\
\includegraphics[width=162pt]{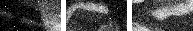}&\hspace{-12pt}
\includegraphics[width=162pt]{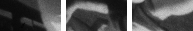}&\hspace{-12pt}
\includegraphics[width=162pt]{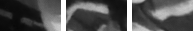}\\
(a)&\hspace{-12pt}(b)&\hspace{-12pt}(c)\\
\includegraphics[width=162pt]{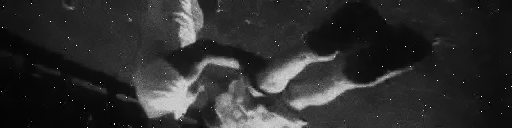}&\hspace{-12pt}
\includegraphics[width=162pt]{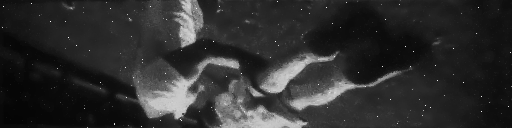}&\hspace{-12pt}
\includegraphics[width=162pt]{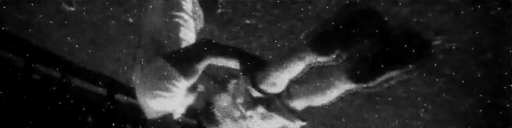}\\
\includegraphics[width=162pt]{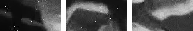}&\hspace{-12pt}
\includegraphics[width=162pt]{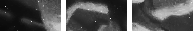}&\hspace{-12pt}
\includegraphics[width=162pt]{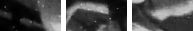}\\
(d)&\hspace{-12pt}(e)&\hspace{-12pt}(f)\\
\end{tabular}
\caption{A representative example from our real test set. (a) A 4-bit frame from the input sequence. (b) Corresponding high-bit resolution image. Frame from recovered video using (c) 3DCNNR, (d) \cite{Maggioni}, (e) \cite{sutour2014adaptive} and (f) \cite{Xue2017VideoEW}.\label{fig:rep}}
\end{center}
\end{figure*}

\begin{table}[htb]
	\renewcommand{\arraystretch}{1.1}
	\caption{Quantitative restoration performance comparison on the real SPAD test dataset.}
	\label{table_evaluation}
	\begin{center}
	\resizebox{\columnwidth}{!}{%
		\begin{tabular}{|c|| c|c|c|c|}
			\hline
			

{\bfseries Measure} & {\cite{Xue2017VideoEW}} & {VBM4D \cite{Maggioni}} &{\cite{sutour2014adaptive}} & 3DCNNR\\
			\hline

			\hline
			\ PSNR & 28.53 & 30.51 & 31.41 &35.54 \\
			\hline
			\ SSIM & 0.7328 & 0.7816& 0.8218 & 0.909\\
			
			\hline
		\end{tabular}
		}
	\end{center}
\end{table}

\noindent \textbf{Real experiments}
We initially show an example of 1-bit real sequence. Fig. \ref{fig:b1_blade} shows a scene wherein the SwissSPAD camera was placed close to a rotating tool. In this particular setup, because of limitations in the data-transfer rate, 1-bit frames were captured at the rate of 42 kfps. A binary image from the input sequence is shown in Fig. \ref{fig:b1_blade} (a). Its corresponding output is shown in Fig. \ref{fig:b1_blade} (b). For comparison, we averaged 15 binary frames to generate a 4-bit sequence. This 4-bit sequence was fed as input to our 3DCNN4B network. In the videos included as supplementary material, one can clearly see the loss of temporal resolution in the output of 3DCNN4B when the averaged sequence was input.

Fig. \ref{fig:b2} shows another rotating tool present in a static background. The input in this scenario was a 2-bit sequence captured at about 25 kfps. While Fig. \ref{fig:b2} (a) shows an input frame, Fig. \ref{fig:b2} (b) shows the corresponding result from 3DCNN2B. By averaging five frames of this sequence, one can obtain a 4-bit sequence. Figs. \ref{fig:b2} (c) and (d) show a 4-bit input and a resultant (3DCNN4B) frame, respectively. We observe that in the static regions, the output of 3DCNN2B does come close to that of 3DCNN4B. The supplementary material contains a comparison of our output with a high-intensity resolution reference image captured when the scene was still.

 We next show results on additional oscilloscope sequences of \cite{burri2014architecture} (Fig. \ref{fig:osci2b4b}). The 2-bit sequence was fed as input to 3DCNN2B, and 4-bit sequence was input to 3DCNN4B. One can observe that the quality of our 2-bit output does come close to that of 4-bit and even with the high-bit observation (Fig. \ref{fig:osci2b4b} (f)) at static regions. This shows that our algorithm is capable of producing high-quality outputs from only 2-bit frames. The output from \cite{sutour2014adaptive} on the 2-bit sequence (Fig. \ref{fig:osci2b4b} (c)) looks quite inferior to our output. The supplementary material contains complete videos and comparisons. We have also shown the result from \cite{s16111961} and compared it with our method.

Subsequently, we compare the performances of 3DCNN1B, 3DCNN2B and 3DCNN3B on the same scene. The image sequence corresponds to breaking of glass with a resolution chart in the background. The scene was captured at 156kfps and at 1-bit resolution. We divided the sequence into groups of seven frames. For 1-bit observations, we keep the central frame in each group and drop six other frames. For 2-bit observations, we average the middle three frames and drop the rest. For 3-bit observations, we average all the seven frames in the group. Essentially, we arrive at 1-bit, 2-bit and 3-bit sequences with similar temporal variations and corresponding to about 22kfps. With these inputs, we obtain the outputs from our restoration scheme. Fig. \ref{fig:glass} shows inputs and outputs at two different instants of time. In the second instant, we see that the particles have been scattered after the breaking of glass. Despite the inputs being highly noisy, we see that the structural information has been recovered quite well. This shows that even with just one-bit measurements, our network model is able to reconstruct scene information quite robustly.

\noindent \textbf{4-bit sequences}
 We quantitatively evaluate different video denoising methods using our real SPAD dataset that has both $4$-bit noisy sequence and the corresponding high-bit sequence. We evaluate the performance on ten randomly selected test-image sequences. The performance of different schemes are presented in Table \ref{table_evaluation}. The algorithms of \cite{Maggioni} and \cite{sutour2014adaptive}, do not handle hot pixels. For these methods, to evaluate score, we replace the intensity of a hot-pixel by the value equal to the median of the $3{\times}3$ neighborhood of that hot pixel (excluding the damaged pixels while calculating median). The other algorithms can handle outliers and this step is not necessary. The table shows that our scheme of 3DCNNR clearly outperforms other methods. A representative example from the SwissSPAD dataset is shown in Fig. \ref{fig:rep}. On close observation of different regions, we can see that the reconstruction is more faithful in the proposed CNN model.
 
We next show a result on a 4-bit sequence captured at 12 kfps. In Fig. \ref{fig:balloon}, show sample frames from the resultant sequences. From the videos, one can clearly observe that the proposed method performs quite well. There has been an improvement in the performance of video denoising schemes when compared to the 1-bit and 2-bit scenarios. However, in the results of \cite{Maggioni}, \cite{sutour2014adaptive} and \cite{Xue2017VideoEW} (supplementary material), we do observe artifacts clearly in regions where there is no motion. In the supplementary material, we show additional results and also include a quantitative evaluation of 3DCNN3B on a synthetic dataset.


\section{Conclusions}
We proposed a video recovery scheme for single-photon counting cameras with sparse photon counts. The performance of our model is quite good in real scenarios despite the training on simulated data. The level of performance achieved by our method in 1-bit and 2-bit scenarios is not possible with any existing approach. Even for 3-bit and 4-bit scenarios, our method outperforms existing video denoising schemes.

In other applications of SPAD cameras such as range imaging and time-of-flight imaging, the number photon counts is of much higher magnitude than the numbers seen in this paper. Our work could serve as a template for developing more photon-efficient techniques to perform these tasks. We would explore this direction in future.

\ifpeerreview
\else


\fi

\bibliographystyle{IEEEtran}
\bibliography{egbib}

\end{document}